\documentclass{vgtc}                          
\ifpdf
  \pdfoutput=1\relax                   
  \usepackage{graphicx}                
  \DeclareGraphicsExtensions{.pdf,.png,.jpg,.jpeg} 
\else
  \usepackage{graphicx}                
  \DeclareGraphicsExtensions{.eps}     
\fi%

\graphicspath{{figures/}{pictures/}{images/}{./}} 

\usepackage{microtype}                 
\PassOptionsToPackage{warn}{textcomp}  
\usepackage{textcomp}                  
\usepackage{mathptmx}                  
\usepackage{times}                     
\usepackage{cite}                      
\usepackage{tabu}                      
\usepackage{booktabs}                  

\usepackage{graphicx}
\usepackage{subfigure}
\usepackage{amsmath,amssymb} 
\usepackage{color}

\onlineid{0}

\vgtccategory{Research}

\vgtcinsertpkg

\title{3D model silhouette-based tracking in depth images for puppet suit dynamic video-mapping}

\author{Guillaume Caron\thanks{e-mail:guillaume.caron@u-picardie.fr} %
\and Mounya Belghiti\thanks{e-mail:belghiti.mounya@gmail.com} %
\and Anthony Dessaux\thanks{e-mail:anthonydessaux@hotmail.com}}
\affiliation{\scriptsize Universite de Picardie Jules Verne, MIS laboratory, Amiens, FRANCE}

\abstract{Video-mapping is the process of coherent video-projection of images, animations or movies on static objects or buildings for shows. This paper focuses on the dynamic video-mapping of the suit of a puppet being moved by its puppeteer on the theater stage. This may allow changing the costume dynamically and simulate light interaction and more. 

Contrary to common video-mapping, the image warping cannot be done once, offline, before the show. It must be done in real-time, and considering a non-flat projection surface, so that the video-projected suit always maps perfectly the puppet, automatically. 

Hence, we propose a new visual tracking method of articulated object, for the puppet tracking, exploiting the silhouette of a 3D model of it, in the depth images of a Kinect v2. Then, considering the precise calibration between the latter and the video-projector, that we propose, coherent dynamic video-mapping is made possible as the presented results show.%
} 

\begin{document}


\firstsection{Introduction}

\maketitle


Video-mapping is well known for shows as images, animations or movies are video-projected on buildings or objects in many cities or theme parks. Contrary to the latter common video-mapping, that is done on static objects, the dynamic video-mapping, \textit{i.e.} the coherent video-mapping of images or animations over moving objects, is still of research interest. The bottleneck is, thus, to track the object on which the video-projector must project coherently the virtual object to augment the reality. 

To deal with that issue, existing works generally consider markers put on the object to track in conjunction with motion tracking systems, sometimes very expensive as for the Face Tracking project\footnote{projection-mapping.org/omote/} of the OMOTE team in which a Vicon system is used. However, such kind of marker-based technique may deal with deforming object shapes~\cite{Fujimoto2014}, at the price of needing to equip the object to track with a set of markers arranged smartly, depending on the object itself. The latter approach, even reaching precise results while considering low-cost RGBD (color and depth) camera, needs a long a priori work to setup. The video-projector, then, displays warped images on the soft object and black color out of it. The object must, then, be moved in the field of the video-projector that is static. This is generally the way followed, even for marker-less techniques, considering contours in 2D IR imaging~\cite{Hashimoto2017}, point clouds with RGBD imaging~\cite{Zhou2016} or keypoints, in 2D RGB imaging, got from the videoprojection over the object to track itself~\cite{Resch2016}, limiting the dynamic video-mapping to highly textured object video-projection, and even with multiple video-projectors~\cite{Siegl2015}. There also exist some adaptive field of view video-projection system with multi-reflections on orientable mirrors in order to display only on the particularly considered object~\cite{Sueishi2015}. Combined with a high speed camera-based tracking system, the latter work is very reactive and deals with highly dynamic moving objects, however constraining the shape of the object to the field of projection shape as circular or spherical targets. \cite{Siegl2015} manages to track the object of interest of a shape that is not constrained by the projector field (the statue of a head) thanks to projective iterative closest point, applied to 3D point clouds got from successive RGBD images, at the price of a GPU implementation to overcome the algorithm complexity bottleneck.

\begin{figure}[!b]
\begin{center}
	\subfigure[]{
		\label{fig:realSuperLycra}
		\includegraphics[height=4cm]{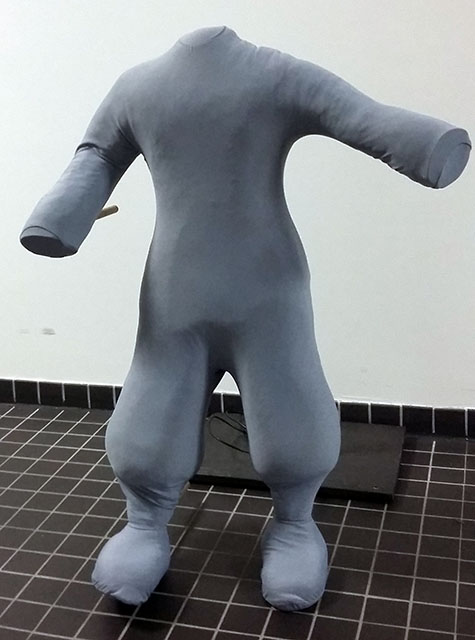}
	}
	\subfigure[]{
		\label{fig:virtualSuperLycra}
		\includegraphics[height=4cm]{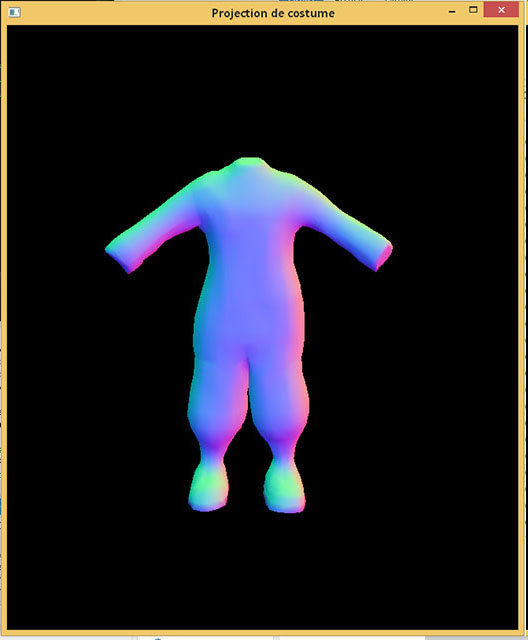}
	}
\caption{(a) The puppet considered for the show, \textit{i.e.} the ``screen'' for videoprojection. (b) The 3D model (mesh) corresponding to the actual puppet.}
\label{fig:SuperLycra}
\end{center}
\end{figure}

None of the previous works in the state-of-the-art tackles the marker-less dynamic video-mapping on a moving and articulated object and it is exactly the purpose of the current paper. Indeed, the goal of the current paper is to propose a new dynamic video-mapping approach that is computationally efficient to interact in real-time with an articulated puppet on which the video-projector displays its suit. The single video-projector situation is considered. The puppet is designed to correspond to the puppet theater play for which this dynamic video-mapping work is led. In this context, the puppet must ``wear'' four different suits during the show with sudden changes and animations on each suit like sunlight simulation or cracks and blood flowing, in order to map the play scenario, and while the puppeteer is acting. The shape of the puppet has been designed by artists to be like a mean shape of every suits. The puppet has no head nor hands to fit the play scenario (Fig.~\ref{fig:realSuperLycra}). Each arm or leg has three degrees of freedom: two joints for each shoulder and ankle and one joint for each elbow and knee. Thus, each arm or leg has three joints as degrees of freedom, leading to a total of twelve joints, in addition to the six degrees of freedom of the puppet 3D pose. 
We propose to consider the depth images of the Kinect v2 to localize the puppet on the theater stage because it brings 3D information that is not disturbed by the light of the video-projector. Indeed, since a lowly textured suit is displayed on the puppet, we cannot consider RGB images. However, even if the Kinect v2 comes with a SDK very efficient to track humans, the lack of head and hands of the puppet makes this existing software failing (no skeleton detected at all on the puppet). Thus, we propose a new tracking approach in the depth images of the puppet moving randomly on the theater stage, exploiting the silhouette of a 3D model of the puppet (Fig.~\ref{fig:virtualSuperLycra}). Considering the silhouette only to recover the 3D pose of the puppet is one of the core contribution of the paper since it can lead to fast processing, even on a laptop. The latter constraint is rather important for the context of the work since the puppet company is often moving from theaters to theaters and considering a Kinect v2 is also relevant for the compactness and cost constraints. 

The rest of the paper is organized as follows. First, the geometrical modeling of the dynamic video-mapping setup is introduced as well as its calibration (Section~\ref{sec:setup}). Then, Section~\ref{sec:tracking} presents the 3D model silhouette-based tracking. Finally, results are shown (Sec.~\ref{sec:results}) before discussion (Sec.~ \ref{sec:discussion}) and the conclusion (Sec.~ \ref{sec:conclusion}).

\section{Video-mapping setup modeling and calibration}
\label{sec:setup}

\subsection{Geometrical modeling}
\label{sec:setupModeling}
 
The dynamic video-mapping setup is geometrically described by four frames, of which three are fixed, \textit{i.e.} $F_p$, $F_d$ and $F_c$, respectively the video projector frame, the depth camera frame and the color camera frame. Both latter frames are in the Kinect v2. The last frame $F_o$ is associated to the puppet, thus mobile. Then, geometrically, the set of extrinsic parameters of the latter setup is made of the following frame changes:
\begin{itemize}
	\item ${^{c}{\bf M}}_d$: the frame change from $F_d$ to $F_c$
	\item ${^{p}{\bf M}}_c$: the frame change from $F_c$ to $F_p$
	\item ${^{d}{\bf M}}_o$: the frame change from $F_o$ to $F_d$
\end{itemize}
Both former frame changes are constant since the Kinect v2 is considered static as well as the video projector. Thus, the puppet tracking will consist in computing ${^{d}{\bf M}}_o$ in real-time.

Finally, intrinsic parameters of Kinect v2 depth and color cameras are also considered as well as for video projector. Assuming the perspective projection fits these three devices, three sets ${^{d}\underline{\pmb{\gamma}}}$, ${^{c}\underline{\pmb{\gamma}}}$ and ${^{p}\underline{\pmb{\gamma}}}$, respectively, of common intrinsic parameters are considered. Each ${^{\cdot}\underline{\pmb{\gamma}}}$ includes four parameters as ${^{\cdot}\alpha}_u$ and ${^{\cdot}\alpha}_v$, for combinations of the focal length and the pixel ratio, and $\left({^{\cdot}u}_0, {^{\cdot}v}_0\right)$ for the principal point, leading to the perspective projection function $pr_{^{\cdot}\underline{\pmb{\gamma}}}\left( \right)$ expression:
\begin{equation}
	{^{\cdot}\underline{\bf u}}
	=
	pr_{^{\cdot}\underline{\pmb{\gamma}}}\left( {^{\cdot}\underline{\bf X}} \right)
	= 
	\begin{pmatrix}
		{^{\cdot}\alpha}_u \frac{{^{\cdot}X}}{{^{\cdot}Z}} + {^{\cdot}u}_0 \vspace{10pt} \\
		{^{\cdot}\alpha}_v \frac{{^{\cdot}Y}}{{^{\cdot}Z}} + {^{\cdot}v}_0
	\end{pmatrix},
\end{equation}
with ${^{\cdot}\underline{\bf X}} = \left({^{\cdot}X}, {^{\cdot}Y}, {^{\cdot}Z}, 1\right)^\mathsf{T}$, a 3D point expressed in frame $F_{\cdot}$, and\linebreak[4]${^{\cdot}\underline{\bf u}} = \left({^{\cdot}u}, {^{\cdot}v}\right)^\mathsf{T}$ its projection in the digital image plane. Superscript $^{\cdot}$ indicates frames $d$, $c$, or $p$.

\subsection{Calibration}
\label{sec:calib}

Two steps are considered for offline calibration: the Kinect v2 calibration itself and the video-projector intrinsic and extrinsic calibration with respect to the Kinect v2 frame. 

\subsubsection{Kinect v2 calibration}
\label{sec:calibKinect}

First, the Kinect v2 is calibrated exploiting the calibration toolbox\footnote{available on request to the authors} of~\cite{Staranowicz2016}, considering 24 pairs of color and depth images of a sphere at various positions. Thus, intrinsic parameters sets ${^{d}\underline{\pmb{\gamma}}}$ and ${^{c}\underline{\pmb{\gamma}}}$, as the extrinsic frame change ${^{c}{\bf M}}_d$, that will, farther in the paper, be represented as a vector, ${^{c}\underline{\bf r}}_d = \left[t_X, t_Y, t_Z, \theta w_X, \theta w_Y, \theta w_Z\right]^\mathsf{T}$, with $t_X$, $t_Y$ and $t_Z$ the translations along the three axes and $\theta$ the rotation angle around the unit axis of rotation of coordinates $\left[w_X, w_Y, w_Z\right]^\mathsf{T}$. Formulas of Rodrigues~\cite{Corke11a} allow to compute ${^{c}\underline{\bf r}}_d$ from ${^{c}{\bf M}}_d$ and ${^{c}{\bf M}}_d$ from ${^{c}\underline{\bf r}}_d$.

\subsubsection{Video-projector calibration}

\begin{figure}[!b]
\begin{center}
	\subfigure[]{
		\label{fig:colorcodedpattern}
		\includegraphics[height=2.5cm]{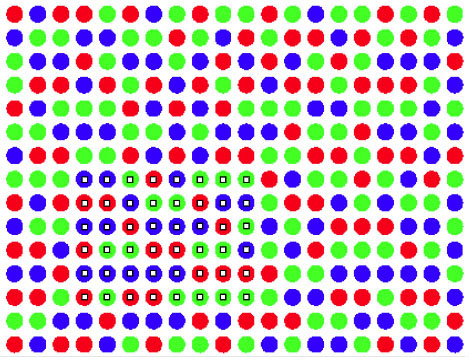}
	}
	\subfigure[]{
		\label{fig:colorcodedpatternmatching}
		\includegraphics[height=2.5cm]{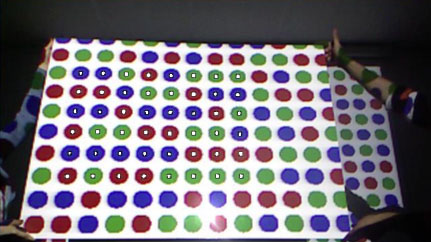}
	}
\caption{(a) Color-coded pattern for automatic matching for video-projector calibration. (b) Automatic matching result for visible dots (white filled squares at the centers of matched dots, corresponding to the ones of (a)).}
\label{fig:colorcoded}
\end{center}
\end{figure}

To compute the video-projector intrinsic parameters ${^{p}\underline{\pmb{\gamma}}}$ and extrinsic ones ${^{p}{\bf M}}_c$, we consider a color-coded calibration pattern made of dots~\cite{Pages2006} in order to make possible the automatic matching between the 2D coordinates ${^{p}\underline{\bf u}}_i^* = \left({^{p}u}_i^*, {^{p}v}_i^*\right)^\mathsf{T}$ of their centers (Fig.~\ref{fig:colorcodedpattern}) and the 3D coordinates ${^{c}\underline{\bf X}}_i = \left({^{c}X}_i, {^{c}Y}_i, {^{c}Z}_i, 1\right)^\mathsf{T}$, expressed in $F_c$, of their observation by the RGBD sensor. Indeed, each 3$\times$3 block of dots is unique in the pattern.

Concretely, a white board is set at various orientations in both video-projector and Kinect v2 fields and several RGBD images are acquired. Then, each color image is considered for automatic dots matching (Fig.~\ref{fig:colorcodedpatternmatching}), and the associated depth image provides the ${^{c}Z}_i$ value corresponding to each ${^{c}\underline{\bf u}}_i = \left({^{c}u}_i, {^{c}v}_i\right)^\mathsf{T}$ dot center in the color image. Then, the full scale 3D coordinates are computed thanks to ${^{c}\underline{\pmb{\gamma}}} = \left\{ {^{c}\alpha}_u, {^{c}\alpha}_v, {^{c}u}_0, {^{c}v}_0\right\}$ and the classical inverse perspective projection equations:
\begin{equation}
\left\{
\begin{matrix}
	{^{c}X}_i = {^{c}Z}_i  \frac{{^{c}u}_i - {^{c}u}_0}{{^{c}\alpha}_u} \vspace{10pt} \\
	{^{c}Y}_i = {^{c}Z}_i  \frac{{^{c}v}_i - {^{c}v}_0}{{^{c}\alpha}_v} \\
\end{matrix}
\right.
.
\label{eq:2D23D}
\end{equation}

Finally, the entire set of 2D-3D correspondences, \textit{i.e.} pairs of ${^{p}\underline{\bf u}}_i^*$ and ${^{c}\underline{\bf X}}_i$, are considered in the same following optimization problem:
\begin{equation}
	\begin{bmatrix}
		{^{p}\hat{\underline{\bf r}}}_c \\
		{^{p}\hat{\underline{\pmb{\gamma}}}}
	\end{bmatrix}
	 = 
	 \mathrm{arg}\min_{{^{p}\underline{\bf r}}_c, {^{p}\underline{\pmb{\gamma}}}} || pr_{^{p}\underline{\pmb{\gamma}}}\left({^{p}\bf M}_c \; {^{c}\underline{\bf X}}_i \right) - {^{p}\underline{\bf u}}_i^* ||
	 .
\end{equation}
The latter is solved by Gauss-Newton optimization as for classical camera calibration, only considering that the 2D digital image points are ``observed'' by the video projector and the corresponding calibration target 3D points are the ones got from the Kinect v2 as explained above. 

\section{Silhouette-based visual tracking}
\label{sec:tracking}

\begin{figure}[!t]
\begin{center}
	\subfigure[]{
		\label{fig:puppetDepthImage}
		\includegraphics[height=3cm]{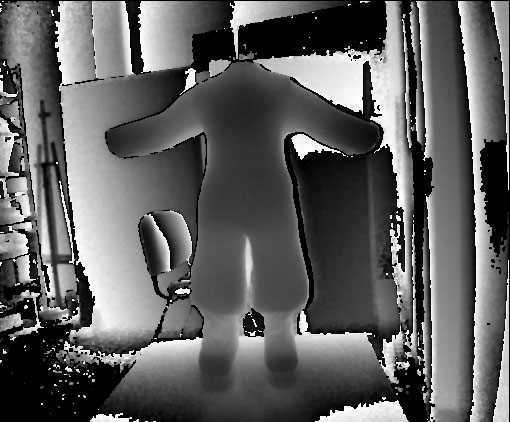}
	}
	\subfigure[]{
		\label{fig:puppetZBuffer}
		\includegraphics[height=3cm]{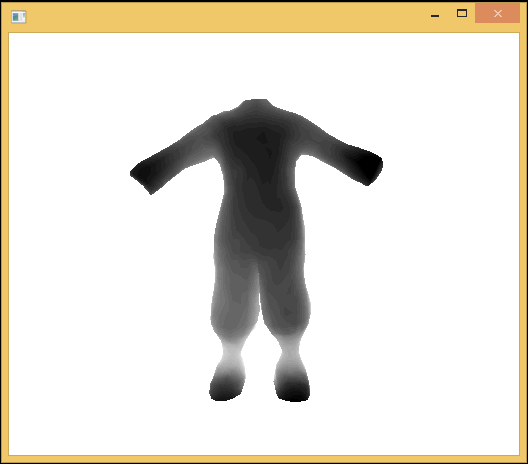}
	}
	
	\subfigure[]{
		\label{fig:puppetSilhouette}
		\includegraphics[height=3cm]{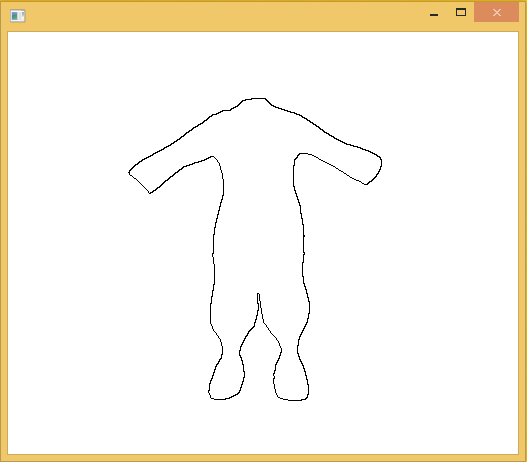}
	}
	\subfigure[]{
		\label{fig:silhouetteOrientations}
		\includegraphics[height=3cm]{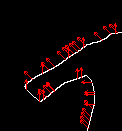}
	}
\caption{(a) A depth image of the puppet from the Kinect v2. (b) A Z-buffer image of the puppet 3D model with the same intrinsic parameters as the actual depth camera. (c) The resulting silhouette pixels computation from (b) (black pixel is a silhouette pixel, white not). (d) zoom on the local orientations of the silhouette.}
\label{fig:puppetDepth}
\end{center}
\end{figure}

The goal is to compute ${^{d}{\bf M}}_o$, \textit{i.e} the puppet pose in the \mbox{Kinect v2} frame, as well as the four arms and legs configurations \mbox{$\underline{\bf q} _m = [q_1, q_2, q_3]^\mathsf{T}$} ($m = 1, 2, 3$ or $4$), from image measurements. The algorithm considers the depth image (Fig.~\ref{fig:puppetDepthImage}) only, to track the puppet and to compute its pose and joints angles, because the dark environment of the theater stage and the video-projection of the puppet suit has not any impact on it, contrary to the RGB image, as theoretically awaited and experimentally observed. 

In this section, we focus on the tracking and pose estimation but not on the detection problem that is solved by imposing to the puppeteer a first puppet pose, approximately fronto-parallel to the Kinect v2 and with the puppet chest approximately aligned with the Kinect v2 main axis. 

Thus, we consider the tracking and pose estimation problems under the family of 3D model-based tracking (see~\cite{Lepetit2005} for a detailed survey). The main idea is to consider an initial pose of the tracked object at which its 3D model can be projected in the image near the observed object. The error between the projected 3D model and the real object is then used for computing pose and joints updates that allow the 3D model to be projected so that it fits perfectly the object. 

For such a problem, the key is to identify the feature adapted to the data that enables matching the reference 3D model describing the actual object in the image of the latter. In the context of this work, we propose to consider the object (puppet) silhouette as the tracking feature. 

\subsection{3D model silhouette computation}
\label{sec:silCom}

The initial silhouette is computed from the 3D rendering of the puppet 3D model. In order to have similar image geometry with respect to the Kinect v2 depth images, the virtual camera parameters are set to the same values as the ones got from the Kinect v2 calibration, ${^{d}\underline{\pmb{\gamma}}}$. Then, a virtual depth image of the puppet 3D model, at initial pose ${^{d^{(0)}}{\bf M}}_o$, is obtained from the Z-Buffer (Fig.~\ref{fig:puppetZBuffer}) of the graphics engine (Ogre 3D\footnote{www.orgre3d.org}, in this work). 

Elementary image processing is applied to detect every Z-buffer image point ${^{d}_{V}\underline{\bf u}_j}$ number $j$ (left subscript $V$ means the 2D coordinates are corresponding to a point in the virtual depth image, \textit{i.e.} the Z-buffer) that is at the border between the puppet body and the background, which is empty (Fig.~\ref{fig:puppetSilhouette}). Concretely, each bone of the virtual puppet is regularly sampled to produce starting points for a search of a high slope of the Z-buffer derivative along the normal axis to the considered bone. The latter method allows to keep the link between each sample point and the puppet bone to which it corresponds. Such information is necessary for the estimation of the articular joint angles (Section~\ref{sec:jointCom}). The local orientation $\phi_S\left({^{d}_{V}\underline{\bf u}_j}\right)$ of the border at every silhouette pixel ${^{d}_{V}\underline{\bf u}_j}$ is also easily got from the latter step (Fig.~\ref{fig:silhouetteOrientations}). 

\subsection{Silhouette samples tracking}
\label{sec:silSamTra}

\begin{figure}[!h]
\begin{center}
	\includegraphics[height=6cm]{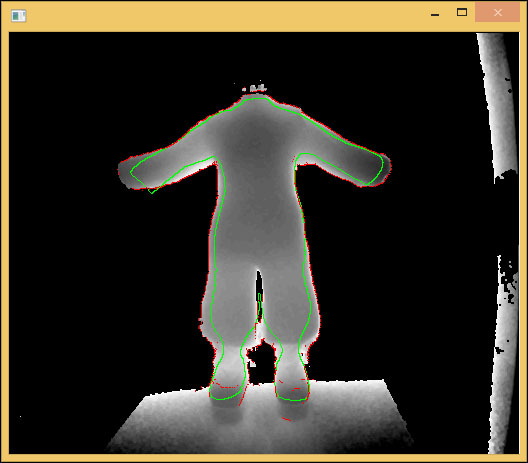}
\caption{Silhouette (green line) sample points tracking: the resulting points are displayed in red. One must note that the range of the depth image is automatically reduced around the depth of the 3D model for more robustness in the image processing.}
\label{fig:movingSilhouette}
\end{center}
\end{figure}

The next step of the algorithm is to look for Kinect v2 depth image points ${^{d}\underline{\bf u}_j}$ corresponding to silhouette sample points ${^{d}_{V}\underline{\bf u}_j}$. Since the virtual camera of the graphics engine considers the same intrinsic parameters as the actual depth camera of the Kinect v2, the ${^{d}_{V}\underline{\bf u}_j}$ 2D coordinates can directly be considered in the actual depth image as starting points in the search for their correspondence. The mobile contour elements algorithm~\cite{Bouthemy1989} applies the latter search along an axis of orientation $\phi_S\left({^{d}_{V}\underline{\bf u}_j}\right)$. The corresponding contour point ${^{d}\underline{\bf u}_j}$ is the one at which the response of the convolution operation between masks of oriented contours and the depth image is maximal. To make more robust the latter search, the depth range of the depth image is adapted to surround the puppet 3D model initial pose, including a safety margin, to avoid clipping the puppet data itself. Figure~\ref{fig:movingSilhouette} shows a silhouette sample points tracking result.

\subsection{Pose computation}
\label{sec:posCom}

Corresponding silhouette 2D points ${^{d}_{V}\underline{\bf u}_j}$ and ${^{d}\underline{\bf u}_j}$ are defined in depth images, the virtual one for the former and the actual one for the latter. Hence, using a similar relationship as Equation~\eqref{eq:2D23D}, for intrinsic parameters ${^{d}\underline{\pmb{\gamma}}}$, we get two sets of corresponding points $\{{^{d^{(0)}}\underline{\bf X}_j}\}$ and $\{{^{d^{(*)}}\underline{\bf X}_j}\}$, describing, respectively, silhouette 3D points at the initial pose ${^{d^{(0)}}{\bf M}}_{o}$, and their corresponding silhouette 3D points at the desired pose ${^{d^{(*)}}{\bf M}}_{o}$, that is unknown. Hence, the next step of the silhouette-based tracking algorithm is to compute the relative transformation ${^{d^{(*)}}{\bf M}}_{d^{(0)}}$.

Every pair ${^{d^{(0)}}\underline{\bf X}_j}$ and ${^{d^{(*)}}\underline{\bf X}_j}$ are linearly related by ${^{d^{(*)}}{\bf M}}_{d^{(0)}}$:
\begin{equation}
	{^{d^{(*)}}\underline{\bf X}_j} = {^{d^{(*)}}{\bf M}}_{d^{(0)}} {^{d^{(0)}}\underline{\bf X}_j},
\end{equation}
and considering the entire set of 3D correspondences, ${^{d^{(*)}}{\bf M}}_{d^{(0)}}$ can directly be computed using a standard rigid transformation estimation algorithm~\cite{Forsyth2002}. 

The latter one produces a precise frame change, in the sense of mean squares, if the data is not noisy and free of outliers. Practically, it is inherent to the silhouette samples tracking approach described in Section~\ref{sec:silSamTra} that the precision of these samples is limited, due to limited depth image resolution and imprecision of measured depths. Furthermore, some outlier samples usually exist if the puppet is partially occluded, for instance. To deal with those issues, the linearly computed ${^{d^{(*)}}{\bf M}}_{d^{(0)}}$ is considered as a good initial guess of the puppet pose for a non-linear optimization scheme, that includes a weighting function to deal with outliers. Thus, considering ${^{d^{(*)}}\underline{\bf r}}_{d^{(0)}}$, the vector representation of pose  ${^{d^{(*)}}{\bf M}}_{d^{(0)}}$, and $\rho(.)$ a robust weighting function, the robust pose optimization problem is written as:
\begin{equation}
	{^{d^{(*)}}\hat{\underline{\bf r}}}_{d^{(0)}}
	 = 
	 \mathrm{arg}\min_{{^{d^{(*)}}\underline{\bf r}}_{d^{(0)}}} \sum_k \; \rho\left( e_k \right)
	 ,
\label{eq:optRobust}
\end{equation}
with $e_k$ the $k$-th element of the error vector $\underline{\bf e}$, made as the stacking of every 3D point-to-point error, \textit{i.e.}:
\begin{equation}
	\underline{\bf e}_{k:k+2}
	 = 
	{^{d^{(*)}}{\bf M}}_{d^{(0)}}{^{d^{(0)}}\underline{\bf X}_j} - {^{d^{(*)}}\underline{\bf X}_j}
	, \; k = 3j.
\end{equation}

To solve the optimization problem of Equation~\ref{eq:optRobust}, we considered the robust iterative closest point non-linear optimization method of~\cite{Fantoni2012}, without the point matching stage, since, in our work, it is already done as described in Section~\ref{sec:silSamTra}. Then, the $\rho(.)$ weighting function is a Huber one~\cite{Huber1964}, weighting less uncertain measures based on the mean and standard deviation of the error vector \underline{\bf e}.

Finally, ${^{d}{\bf M}}_o = {^{d^{(*)}}{\bf M}}_{o}$ is obtained by the usual transformation composition:
\begin{equation}
	{^{d}{\bf M}}_o = {^{d^{(*)}}{\bf M}}_{d^{(0)}} \;{^{d^{(0)}}{\bf M}}_{o}.
	\label{eq:dMo}
\end{equation}

\subsection{Joints computation}
\label{sec:jointCom}

This section focuses on the estimation of joint angles $\underline{\bf q} _m$ of an arm of the puppet ($m$ = 1), with two orthogonal joints ($q_1$ and $q_2$) at the shoulder and one joint ($q_3$) at the elbow, but it is straightforward to apply the same method for the second arm and the legs.

First of all, the arm geometry is got from the standard representation of Denavit-Hartenberg~\cite{Corke11a}, based on relative orientation of joints axes and arm segments length. Then, thanks to ${^{d}{\bf M}}_o$ (eq.~\ref{eq:dMo}) and the rigid transformation from $F_o$ to the arm basis frame $F_m$, \textit{i.e.} ${^{o}{\bf M}}_m$, each 3D silhouette points pair is transformed to be expressed in the arm basis frame as ${^m\underline{\bf X}_j}$, for the initial (got from the virtual depth image) and ${^m\underline{\bf X}_j^*}$, for the desired (got from the Kinect depth image). Only 3D points associated to the bones $B_1$ and $B_2$ of the arm $m$ (Section~\ref{sec:silCom}) are considered to estimate the arm configuration $\underline{\bf q}_m^*$, that is unknown, corresponding to silhouette points ${^m\underline{\bf X}_j^*}$, starting from the known initial configuration $\underline{\bf q}_m$, corresponding to points ${^m\underline{\bf X}_j}$. Then, it is straightforward to express the later points in bone frames $F_{B_1}$ and $F_{B_2}$ to which they correspond, by computing ${^{m}{\bf M}_{B_1}}$ and ${^{m}{\bf M}_{B_2}}$ from $\underline{\bf q}_m$, as ${^{B_b}\underline{\bf X}_j}$ ($b = 1$ or $2$), thanks to the forward geometrical model of the arm~\cite{Corke11a}.

Thus, the $\underline{\bf q}_m^*$ estimation is expressed as the following robust optimization problem:
\begin{equation}
	\hat{\underline{\bf q}_m^*} = \mathrm{arg}\min_{\underline{\bf q}} \sum_j \rho \left( {^{m}\underline{\bf X}_j} - {^{m}\underline{\bf X}_j^*} \right)
	,
	\label{eq:coutq}
\end{equation}
where ${^{m}\underline{\bf X}_j}$ depends on frame changes ${^{m}{\bf M}_{B_1}}$ and ${^{m}{\bf M}_{B_2}}$, themselves depending on joint variables $\underline{\bf q}_m$. Thus, more precisely:
\begin{equation}
	{^{m}\underline{\bf X}_j} = {^{m}{\bf M}_{B_b}} (\underline{\bf q}_m) \; {^{B_b}\underline{\bf X}_j}
	,
\end{equation}
knowing that there is not any segment between joints 1 and 2, so ${^{m}{\bf M}_{B_1}}$ depends on $q_1$ and $q_2$ and ${^{m}{\bf M}_{B_2}}$ depends on $q_1$, $q_2$ et $q_3$. 

Equation~\ref{eq:coutq} is solved by non-linear optimization using the Gauss-Newton gradient descent-like method, of which the core issue is to express the Jacobian matrix of ${^{m}\underline{\bf r}_{B_b}}$, the vector representation of ${^{m}{\bf M}_{B_b}}$, over $\underline{\bf q}_m$, that is very common in robotics~\cite{Paul1978}. Each optimization iteration $it$ leads to a arm configuration update $\dot{\underline{\bf q}}_m$ so that:
\begin{equation}
	\underline{\bf q}_m^{it} = \underline{\bf q}_m^{it-1} + \dot{\underline{\bf q}}_m.
\end{equation}
Then, ${^{m}\underline{\bf X}_j}$ are updated too thanks to new ${^{m}{\bf M}_{B_b}}$ got from $\underline{\bf q}_m^it$ and the algorithm loops until the criterion of equation~\ref{eq:coutq} is stable.

\subsection{Overview of the contribution}
\label{sec:oveCon}

To our knowledge, the closest related work~\cite{Petit12b} considers every object contours in a virtual depth image and minimizes the distances between tangent lines to these contours and contour points detected in the RGB image plane, based on geometry and appearance, to optimize non-linearly the object pose. If we applied some similar image processing (Section~\ref{sec:silSamTra}), the contributions of our approach are in the following items, compared to~\cite{Petit12b}:
\begin{itemize}
	\item using a depth image, contrary to a RGB image (no appearance)
	\item considering the silhouette only (no additionnal contour)
	\item image processing leads to two sets of corresponding 3D points
	\item robust registration between 3D-3D correspondences
\end{itemize}

Another work~\cite{Comport2005} tackled the estimation of a robotic arm configuration from the edges of its observation by a 2D camera. In the later work, salient edges characterize the robot shape and these edges are considered to form the wireframe 3D model of the robotic arm. Point-to-edge 3D model-based tracking is implemented with two joints as degrees of freedom. 
If the problematic is very similar to ours, however focusing on a unique arm, the contributions of our approach are in the following items, compared to~\cite{Comport2005}:
\begin{itemize}
	\item considering two arms and two legs of three joints each
	\item considering a possible moving basis
	\item considering the arm silhouette only (no salient edges exists on the puppet)
	\item registration is made in 3D space, from points only
\end{itemize}

Directly considering correspondences of 3D points, leading, thus, to the direct computation of the pose and joints updates in the tracking process, costs less computation power than in 2D, since we avoid their projection in the image plane. However, if the initial pose is not close enough to the optimal one, several iterations of the algorithm described from Section~\ref{sec:silCom} to~\ref{sec:jointCom} might be applied on the same depth image, to reach a higher pose precision, due to the strong assumption that 2D points on the actual puppet contour exactly correspond to the silhouette sample points got from the virtual depth image. In practice, the number of iterations is limited to the depth image acquisition rate to avoid time delay in the video-mapping. Two iterations experimentally reveals sufficient for slow motion of the puppet. 

Finally, none of the contour 3D model-based tracking approaches close to ours~\cite{Petit12b, Comport2005} considers the dynamic video-mapping application. 


\section{Puppet dynamic video-mapping results}
\label{sec:results}

\subsection{Hardware setup and software}

\begin{figure}[!b]
\begin{center}
	\includegraphics[height=3.75cm]{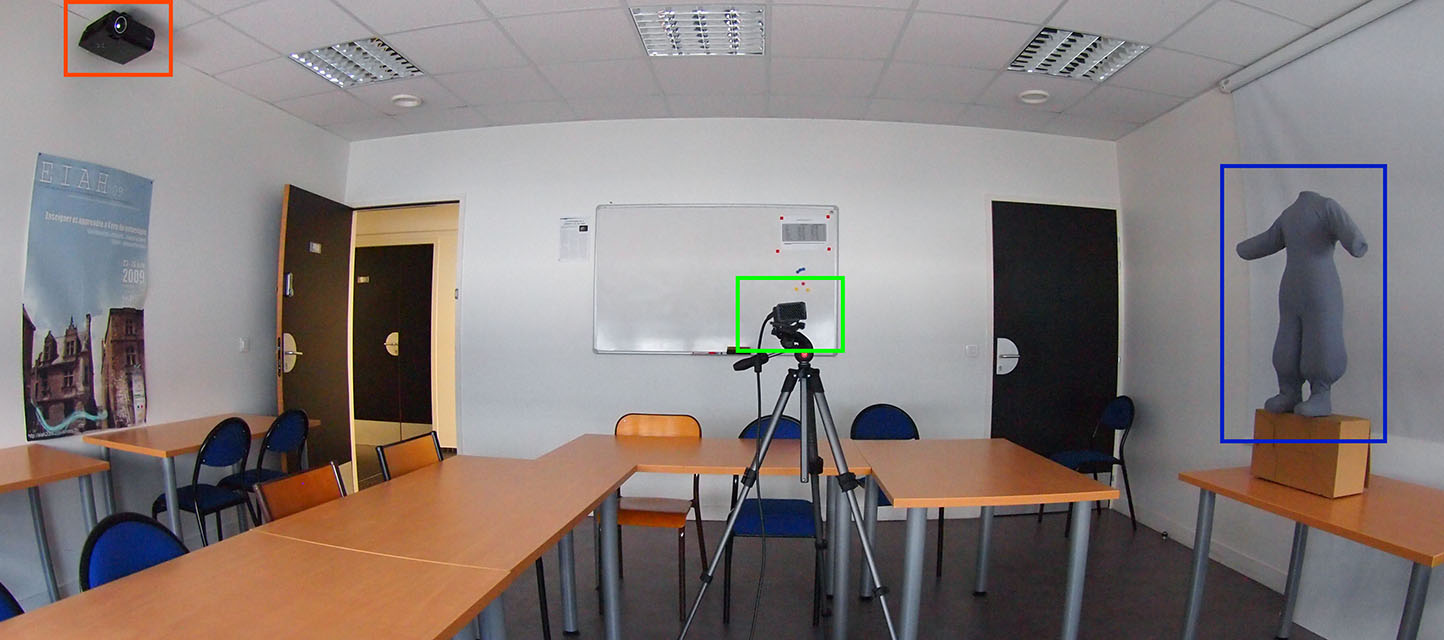}
\caption{Panoramic view of the hardware setup: the Kinect v2 (central green rectangle) is pointed to the puppet (right blue rectangle) on which the video-projector (left red rectangle) must project its suit.}
\label{fig:setup}
\end{center}
\end{figure}

The hardware setup (Fig.~\ref{fig:setup}) is made of:
\begin{itemize}
	\item the puppet itself, that can be moved by the puppeteer in a volume of\linebreak[4]2 m $\times$ 1.5 m $\times$ 1.5 m
	\item a Vivitek D940VX video projector, set to enlighten the puppet on stage
	\item the Kinect v2, facing, at most, the stage volume, while being out of the video projector field
\end{itemize}

To reach the required precision, the Kinect v2 has been calibrated, as introduced in Section~\ref{sec:calibKinect} (Table~\ref{tab:Kinectv2Parameters}).
\begin{table}[!h]
\caption{Kinect v2 intrinsic (top) and extrinsic (bottom) parameters. The pixel reprojection mean error for these parameters is 1.50 (standard deviation: 0.88).}
\begin{center}
Intrinsic parameters (pixel unit)\\
\begin{tabular}{|c|c|c|c|c|c|}
\hline
camera & frame & ${^{\cdot}\alpha}_u$ & ${^{\cdot}\alpha}_v$ & ${^{\cdot}u}_0$ & ${^{\cdot}v}_0$\\
\hline
depth & $F_d$ & 359.90 & 359.21 & 239.80 & 208.67 \\
color & $F_c$ & 1065.92 & 1063.69 & 944.65 & 549.32 \\
\hline
\end{tabular}
\\ \vspace{10pt}
Extrinsic parameters ($t_{.}$ in mm and $\theta w_{.}$ in rad)\\
\begin{tabular}{|c|c|c|c|c|c|c|}
\hline
frame & $t_X$ & $t_Y$ & $t_Z$ & $\theta w_X$ & $\theta w_Y$ & $\theta w_Z$ \\
change & & & & & & \\
\hline
${^{c}{\bf M}}_d$ & -55.64 & 0.95 & 7.04 & -0.02 & -0.01 & -0.00 \\
\hline
\end{tabular}
\end{center}
\label{tab:Kinectv2Parameters}
\end{table}%

Then, the calibration of the video-projector respectively to the Kinect v2 needs image processing and calibration optimization which are done with the ViSP~\cite{Marchand05b} C++ library, even if the automatic matching of dots (Fig.~\ref{fig:colorcoded}) is our implementation of a part of~\cite{Pages2006}. Video-projector calibration results are shown in Table~\ref{tab:VPKcalibration}, only valid for the given setup. Indeed, ${^{p}v}_0$, for instance, is negative because the video-projection field is not equiangular around the optical axis of the video-projector lens. This is needed by the configuration depicted in Figure~\ref{fig:setup}, where the video projector is set on one side of the room ceiling and the usual video-projection surface is on the opposite wall.
\begin{table}[!h]
\caption{Video-projector intrinsic (top) and extrinsic (bottom) parameters obtained for the setup (Fig.~\ref{fig:setup}) considered in the experimental results (Sec.~\ref{sec:results}). 
}
\begin{center}
Intrinsic parameters (pixel unit)\\
\begin{tabular}{|c|c|c|c|c|c|}
\hline
${^{p}\alpha}_u$ & ${^{p}\alpha}_v$ & ${^{p}u}_0$ & ${^{p}v}_0$\\
\hline
2145.99 & 2138.92 & 478.96 & -47.94 \\
\hline
\end{tabular}
\\ \vspace{10pt}
Extrinsic parameters ($t_{.}$ in mm and $\theta w_{.}$ in rad)\\
\begin{tabular}{|c|c|c|c|c|c|c|}
\hline
frame & $t_X$ & $t_Y$ & $t_Z$ & $\theta w_X$ & $\theta w_Y$ & $\theta w_Z$ \\
change & & & & & & \\
\hline
${^{p}{\bf M}}_c$ & 80.19 & 1456.58 & 2612.01 & 0.00 & 0.00 & -0.00 \\
\hline
\end{tabular}

\end{center}
\label{tab:VPKcalibration}
\end{table}%

Finally, the rest of the tracking and mapping process is done online thanks to the Kinect v2 that acquires 30 color and depth images per second and thanks to the video-mapping process that takes, for a unique iteration of the tracking in one depth image acquisition, 13~ms to 17~ms. Thus, when needed, it makes possible to run two iterations of the tracking in one depth image to improve precision. Moving the actual puppet and having the real-time dynamic video-mapping of its suit is, hence, possible. The considered computation hardware is a HP laptop with Intel i7 4800MQ 2.7-3.7 GHz microprocessor, 16 GB of RAM (approx. 150MB used for the dynamic video-mapping) and a NVIDIA Quadro K610M graphics card (only used once or twice per acquired depth image, as stated in Section~\ref{sec:oveCon}). The entire program is written in C++, without particular code optimization, using ViSP and Ogre3D libraries. 

\subsection{Video-mapping}

\begin{figure}[!b]
\begin{center}
	\subfigure[]{
		\label{fig:texturedPuppet}
		\includegraphics[height=4cm]{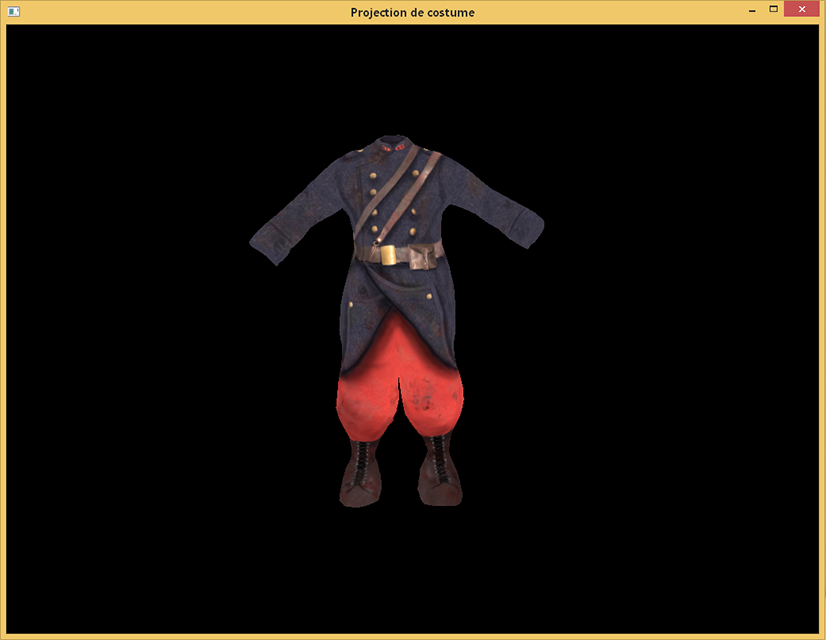}
	}
	\subfigure[]{
		\label{fig:puppetMapping}
		\includegraphics[height=4cm]{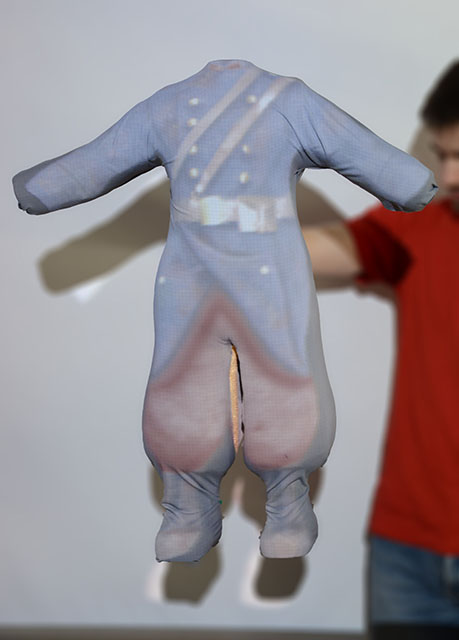}
	}
\caption{(a) The textured 3D model of the puppet. (b) Puppet suit video mapping result (brightness of the video-projector is set to the maximum value to see the projected suit and the actual puppet on the photo).}
\end{center}
\end{figure}

First of all, in order to evaluate the video-mapping precision, a first experiment is made with a rigid version of the puppet, to ease the evaluation process. Then, once the current puppet pose ${^{d}{\bf M}}_o$ is computed from the depth image processing, a color virtual image of the textured puppet can be rendered (Fig.~\ref{fig:texturedPuppet}) from the video-projector point of view $F_p$, thanks to ${^{c}{\bf M}}_d$ and ${^{p}{\bf M}}_c$ frame changes and ${^{p}\underline{\pmb{\gamma}}}$ intrinsic parameters. This virtual image is simply displayed by the video-projector and the puppet virtual suit is mapped on the actual puppet (Fig.~\ref{fig:puppetMapping}).

In such a setup, qualitative evaluation is rather easy since the suit is displayed on the puppet at the correct position and orientation as well as the correct scale, globally (see Fig.~\ref{fig:puppetMapping} where the top, bottom, left and right parts of the suit fits well the puppet). On the other hand, quantitative evaluation is more difficult to obtain since, to our knowledge, there is not any available benchmark or dataset on that topic and ground truth is tough to get. 

A basic mapping precision evaluation can however be led in the photo observing the video-mapping result. Indeed, in the photo of Figure~\ref{fig:puppetMapping}, manual selection of pixels corresponding to the puppet, to the suit projected on the puppet and to the suit projected outside the puppet (on the background), can be done. Then, counting the number of image pixels in these selections may lead to some statistics of correct video-mapping of which a perfect mapping could be defined as ``every pixel of the puppet in the photo is one of the suit \underline{and} not any pixel of the suit is projected out of the puppet''. Pixels counts reported in Table~\ref{tab:quantEval} for the photo of Figure~\ref{fig:puppetMapping} indicates a correct mapping at 90\%, the main source of incorrect mapping being the difference between the 3D model and the actual puppet (80\% of the remaining 10\% since suit parts out of the puppet represent only 2\%). 
\begin{table}[!h]
\caption{Quantitative evaluation of video-mapping correctness (total number of pixels in the considered photo of Figure~\ref{fig:puppetMapping}: 2662 $\times$ 3712).}
\begin{center}
\begin{tabular}{|c|c|c|}
\hline
selection & number of pixels & ratio w.r.t. puppet \\
 &  in the photo  & number of pixels  \\
\hline
puppet & 3 026 107 & 100.00\% \\
suit on the puppet & 2 742 388 & 90.62\% \\
suit outside the puppet & 67 380 & 2.23\% \\
\hline
\end{tabular}
\end{center}
\label{tab:quantEval}
\end{table}%

\subsection{Dynamic video-mapping}

\subsubsection{Rigid puppet}

\begin{figure}[!t]
\begin{center}
	\subfigure[]{
		\includegraphics[width=3.9cm]{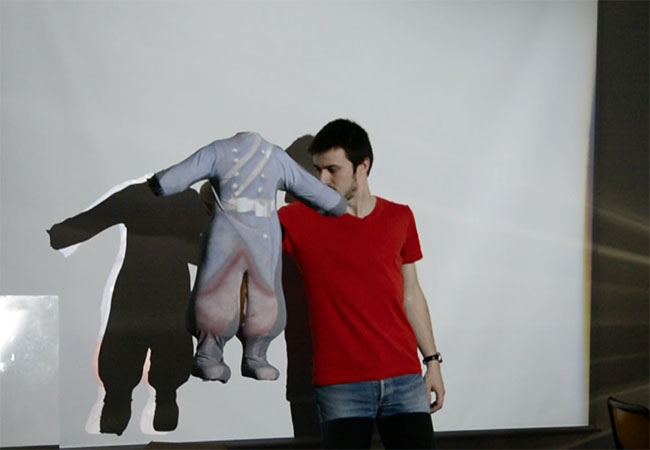}
	}
	\subfigure[]{
		\label{fig:seq_03}
		\includegraphics[width=3.9cm]{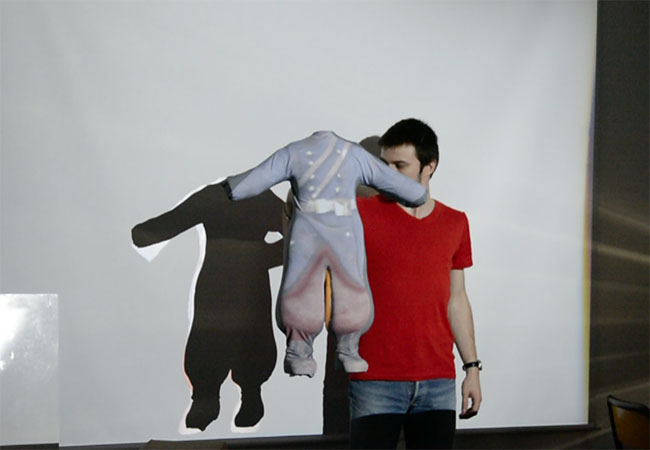}
	}
	
	\subfigure[]{
		\includegraphics[width=3.9cm]{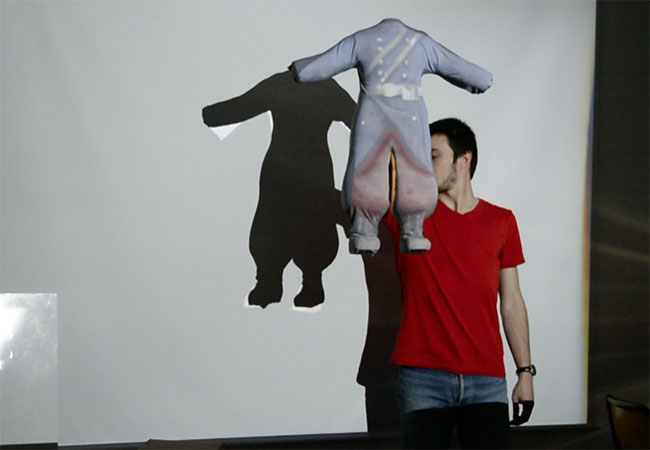}
	}
	\subfigure[]{
		\includegraphics[width=3.9cm]{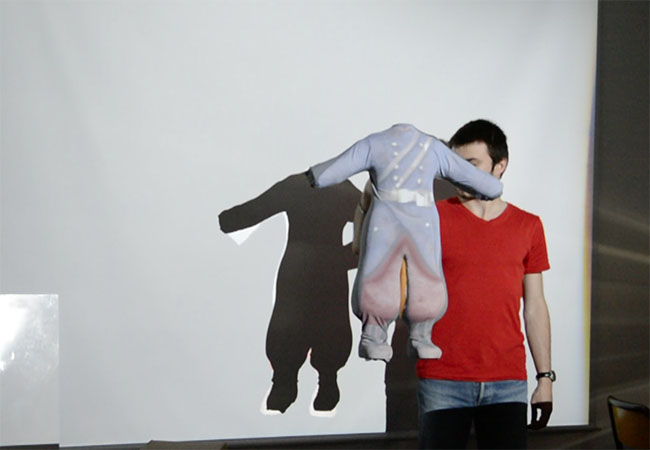}
	}
	
	\subfigure[]{
		\includegraphics[width=3.9cm]{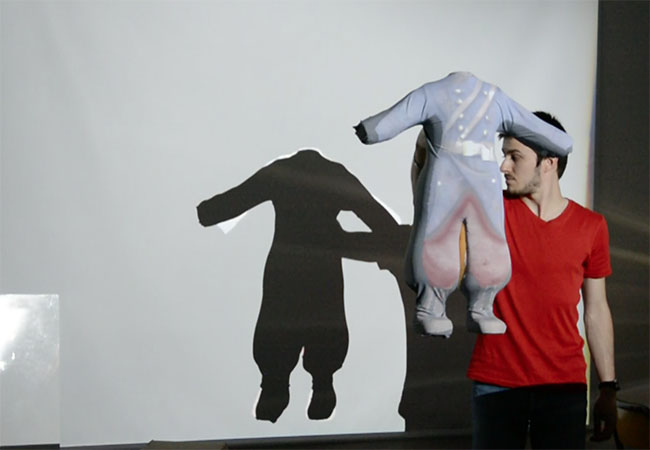}
	}	
	\subfigure[]{
		\label{fig:seq_14}
		\includegraphics[width=3.9cm]{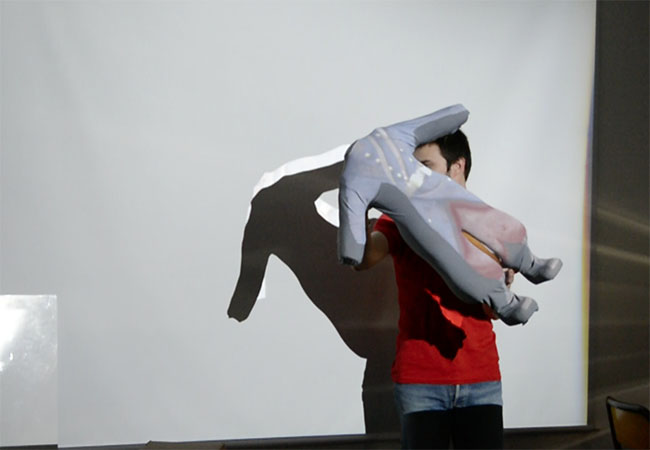}
	}
	
	\subfigure[]{
		\label{fig:seq_15}
		\includegraphics[width=3.9cm]{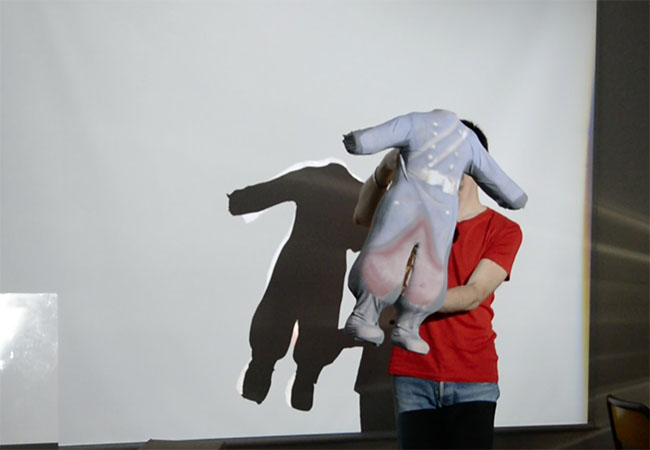}
	}	
	\subfigure[]{
		\includegraphics[width=3.9cm]{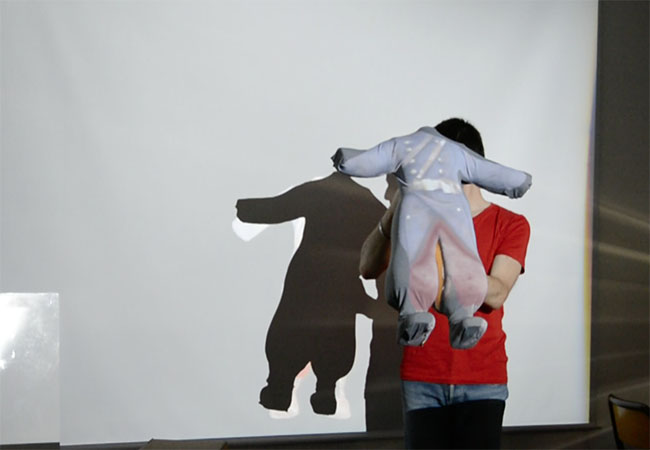}
	}
	
	\subfigure[]{
		\includegraphics[width=3.9cm]{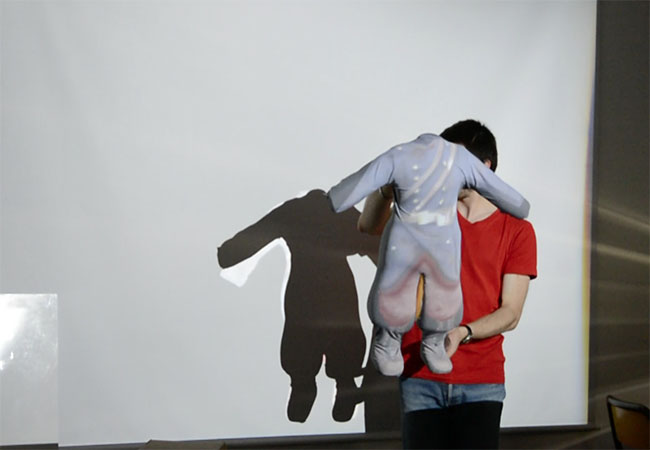}
	}
	\subfigure[]{
		\includegraphics[width=3.9cm]{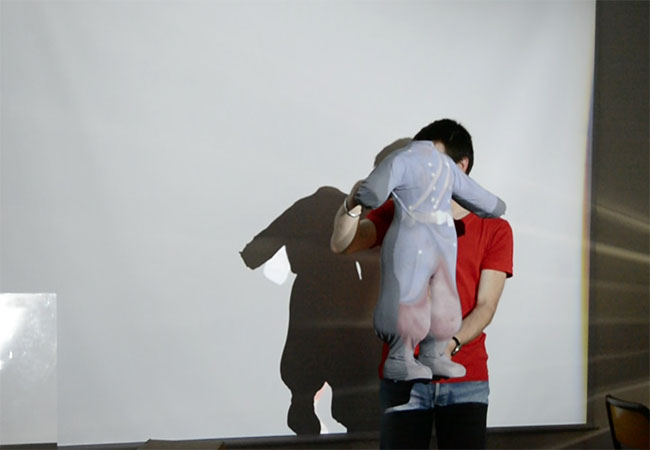}
	}
\caption{Puppet suit dynamic video-mapping images extracted from a single sequence of one minute in which various poses with automatic mapping are demonstrated, filmed with a third party camera.}
\label{fig:dynVidMap}
\end{center}
\end{figure}

Still with the rigid puppet, the tracking and pose estimation approach proposed in this paper is applied. Thus, the puppet is moved in the common field of the Kinect v2 and the video-projector. Figure~\ref{fig:dynVidMap} shows that the suit follows well the puppet for pure translations (Figure~\ref{fig:dynVidMap}(a-e)) as well as for three axes rotations (Figure~\ref{fig:dynVidMap}(f-j)). Obviously, since held by a human, the six degrees of freedom of the puppet pose are mandatory to be considered to ensure a correct tracking and mapping, and all of them are used in this experiment.

\subsubsection{Articulated puppet}

\begin{figure}[!h]
\begin{center}
	\subfigure[]{
		\includegraphics[width=4cm]{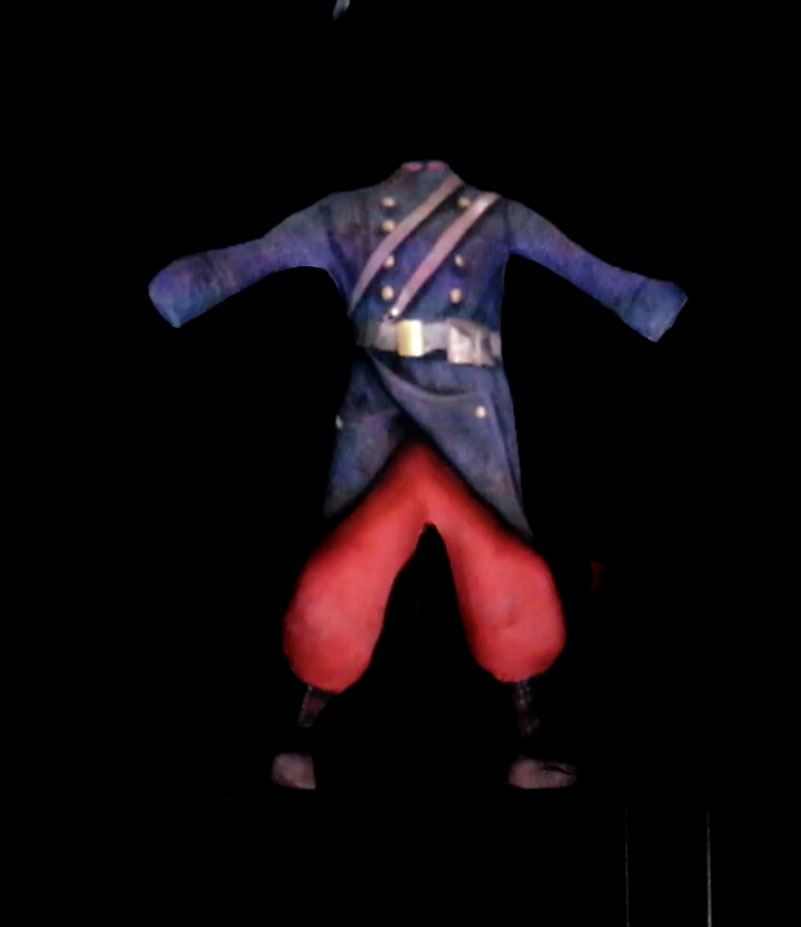}
	}
	\subfigure[]{
		\includegraphics[width=4cm]{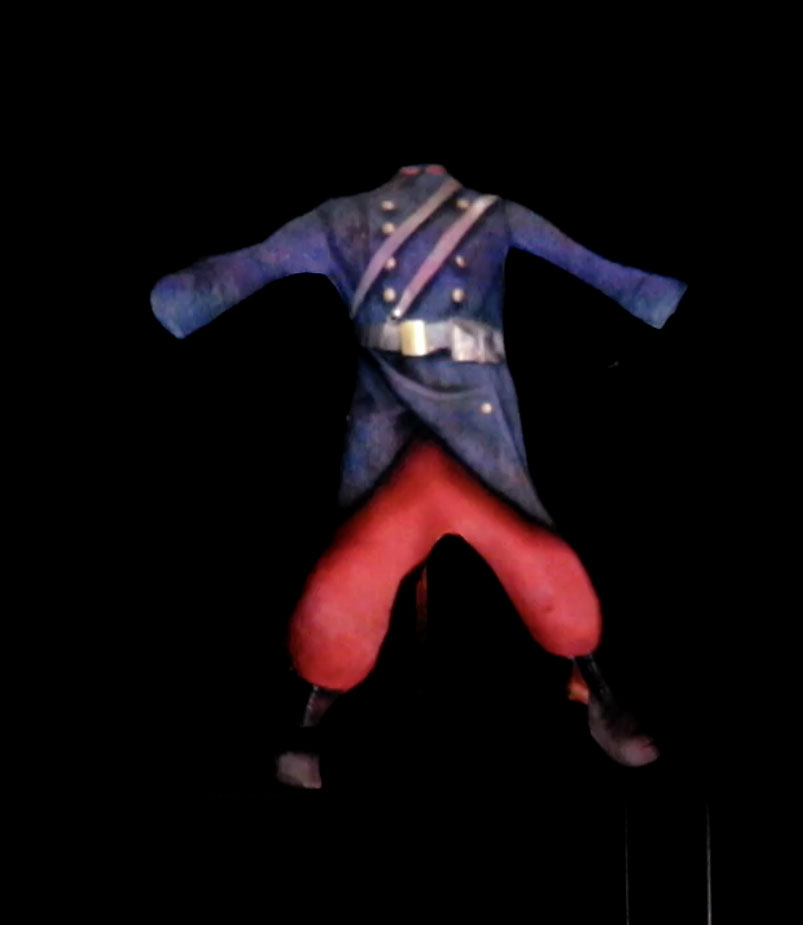}
	}
	
	\subfigure[]{
		\includegraphics[width=4cm]{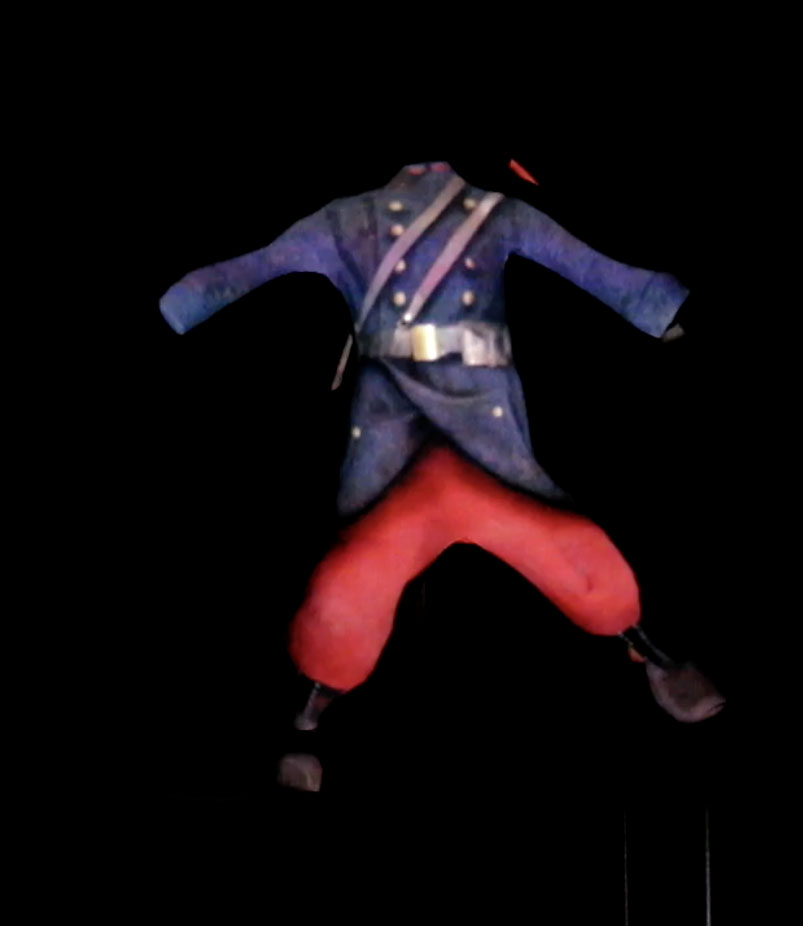}
	}
	\subfigure[]{
		\includegraphics[width=4cm]{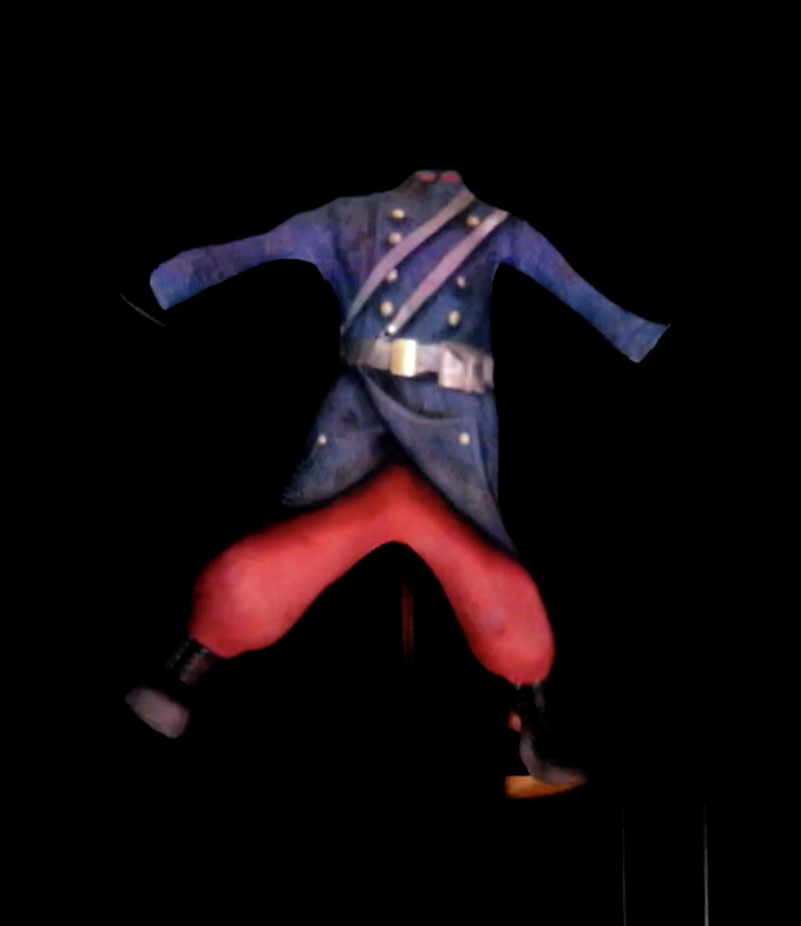}
	}
	
	\subfigure[]{
		\includegraphics[width=4cm]{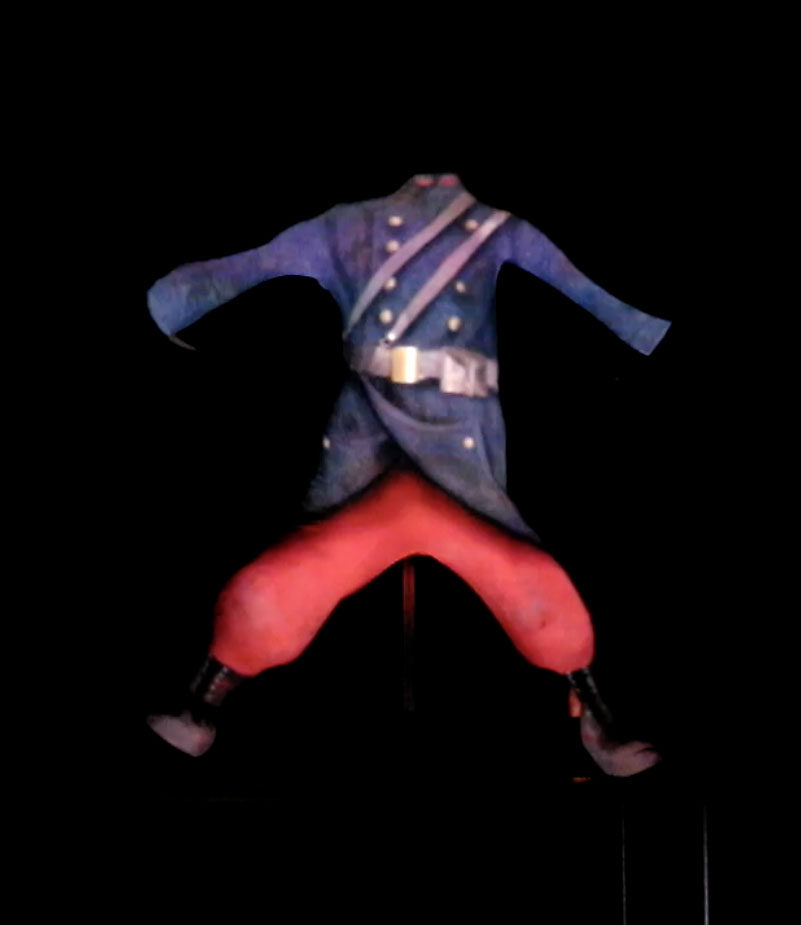}
	}	
	\subfigure[]{
		\label{fig:artDynVidMapF}
		\includegraphics[width=4cm]{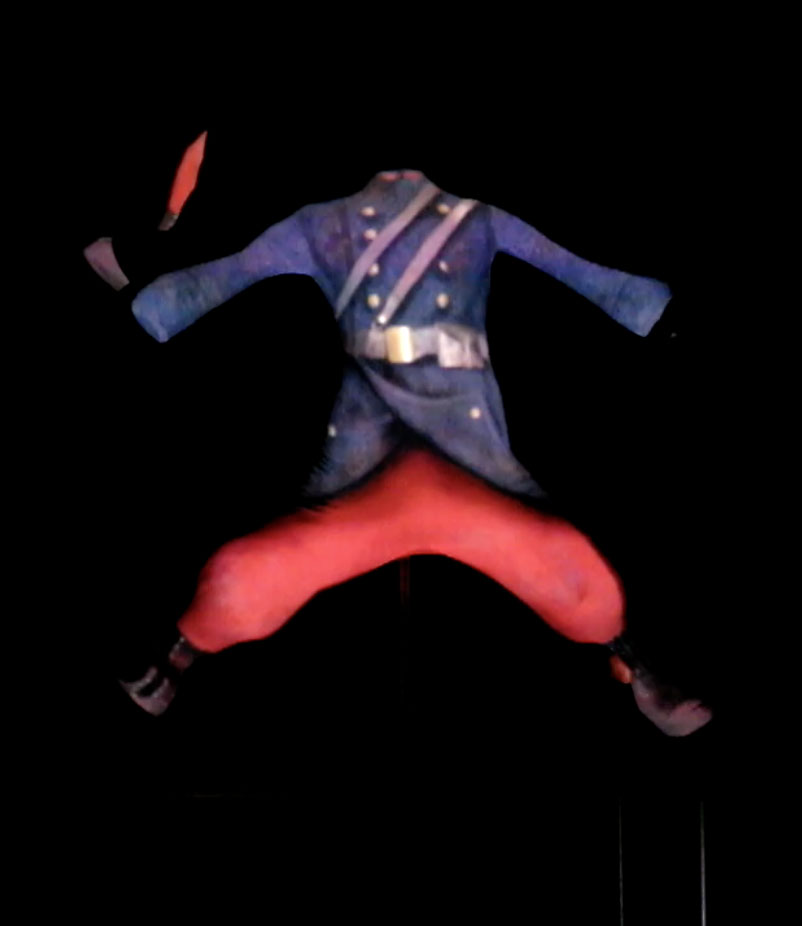}
	}
	
	\caption{Articulated puppet suit dynamic video-mapping images extracted from a single sequence of forty seconds in which various possible legs motion are encountered with correct automatic mapping are demonstrated, filmed with a third party camera. The legs are moved manually by the puppeteer, thanks to wooden sticks at the back of each foot.}
\label{fig:artDynVidMap}
\end{center}
\end{figure}

The third result considers the articulated puppet for which the joints angles estimation is necessary to display correctly the suit arms and legs. Figure~\ref{fig:artDynVidMap} shows both legs are manually moved by the puppeteer with a coherent update of the joint angles. Figure~\ref{fig:artDynVidMap} (a-c) shows the dynamic video-mapping result when moving the puppet left leg. Then, the puppeteer moves it back to its initial state (almost vertical) and moves the other leg (Fig.~\ref{fig:artDynVidMap} (d)). Finally, both legs are actuated simultaneously (Fig.~\ref{fig:artDynVidMap} (e-f)).

\section{Discussion}
\label{sec:discussion}

Some limits are identified through shifts that appear in some images of the presented sequences between the suit and the screen-puppet. It is mainly due to two facts: 
\begin{itemize}
	\item latency induced by devices and processing: indeed, the difference between the time at which the puppet reaches a given pose and the time at which the suit is displayed at that pose is not null (Kinect v2 minimum latency is 20~ms, dynamic video-mapping algorithm minimum processing time is 13~ms and video-projection latency is between 33~ms to 160~ms). 
	\item puppet motion along an axis approximately tangent to a large proportion of the silhouette: indeed, in that case, several mismatched silhouette points are considered in the pose update computation, concretely leading to a low error at that points and an underestimated pose update, needing several iterations of the tracking algorithm to well match the puppet in the depth image.
	\item Considering a black background may not always be relevant since when slight registration error appears, puppet suit parts can be displayed on the background stage (Fig.~\ref{fig:artDynVidMapF}: a part of the red trousers and of the puppet right shoe is displayed on the background stage, leading to undesired visual effects).
\end{itemize}
Both former facts are particularly visible when the puppeteer moves a bit faster (Fig.~\ref{fig:seq_14}). When he slows down the puppet motion, the suit progressively maps again the puppet (Fig.~\ref{fig:seq_15}). 

\section{Conclusion}
\label{sec:conclusion}

This paper has introduced the vision-based tracking of an articulated puppet in the context of dynamic video mapping. The silhouette feature only is considered in the 3D model-based tracking approach, exploiting depth images of an RGBD camera. The proposed calibration procedure as well as the efficience of the puppet tracking, in precision and computational load, leads to precise dynamic video-mapping in real-time, without any dedicated hardware nor GPU implementation (the GPU is only used as for what it is built for: render images of a 3D model). 

Future works will handle the identified limits of the current algorithm as the latency issue due to devices and the algorithm itself for a fast set of motions. A deeper quantitative evaluation will also be led.

\section{Acknowledgements}
The authors wish to highlight the puppeteers company named ``Le Tas de Sable - Ch\'{e}s Panses Vertes'' at the origin of this work: Sylvie Baillon, for directing the puppet show, and Eric Goulouzelle, for making the screen puppets shown in this paper. The puppet 3D model was designed by Alexis Leleu, and textured by Margot Briquet, from the 3D animation school named ``Waide Somme''. 

\bibliographystyle{abbrv-doi}
\bibliography{puppetVideoMapping}


\end{document}